\documentclass[a4paper]{article}%
\usepackage{amsmath}%
\usepackage{amsfonts}%
\usepackage{amssymb}%
\usepackage{graphicx}
\usepackage{fullpage}

\usepackage{math_defs}

\begin{document}

\title{PLDA with Two Sources of Inter-session Variability}
\author{Jes\'{u}s Villalba\\\\
  Communications Technology Group (GTC),\\ Aragon Institute
  for Engineering Research (I3A),\\ University of Zaragoza, Spain\\
  \small \tt villalba@unizar.es}
\date{Nov 6, 2012}
\maketitle

\section{The Model}

\subsection{PLDA}

We take a linear-Gaussian generative model $\model$. We suppose that
we have i-vectors of the same conversations recorded simultaneously by
different channels or different noisy conditions. Then, an.
i-vector $\phivec_{ijk}$ of speaker $i$, session $j$ recorded in
a channel $l$ can be written as:
\begin{equation}
  \phivec_{ijl}=\muvec+\Vmat\yvec_{i}+\Umat\xvec_{ij}+\epsilon_{ijl}
  \label{eq:2cov_model}
\end{equation} 
where $\muvec$ is speaker independent term, $\Vmat$ is the eigenvoices
matrix, $\yvec_i$ is the speaker factor vector, $\Umat$ is an the
eigenchannels matrix, $\xvec_{ij}$ and $\epsilon_{ijl}$ is a channel
offset. The term $\xvec_{ij}$ must be the same for all the recordings
of the same conversation. The term $\epsilon_{ijl}$ accounts for the
channel variability.

We assume the following priors for the variables:
\begin{align}
  \label{eq:plda_yprior}
  \yvec&\sim\Gauss{\yvec}{\zerovec}{\Imat} \\
  \label{eq:plda_xprior}
  \xvec&\sim\Gauss{\xvec}{\zerovec}{\Imat} \\
  \label{eq:plda_eprior}
  \epsilon&\sim\Gauss{\epsilon}{\zerovec}{\iWmat}
\end{align}
where $\mathcal{N}$ denotes a Gaussian distribution; and $\Dmat$
is a full rank precision matrix. $\phivec$ is an observable
variable and $\yvec$ and $\xvec$ are hidden variables.

\subsection{Notation}

We are going to introduce some notation:
\begin{itemize}
\item Let $\Phimatd$ be the development i-vectors dataset.
\item Let $\Phimatt=\left\{l,r\right\}$ be the test i-vectors.
\item Let $\Phimat$ be any of the previous datasets.
\item Let $\thetad$ be the labelling of the development dataset. It
  partitions the $N_d$ i-vectors into $M_d$ speakers. Each speaker has
  $H_i$ sessions and each session can be recorded by $L_{ij}$
  different channels. 
\item Let $\thetat$ be the labelling of the test set, so that $\thetat
  \in \left\{\tar,\nontar\right\}$, where $\tar$ is the hypothesis
  that $l$ and $r$ belong to the same speaker and $\nontar$ is the
  hypothesis that they belong to different speakers.
\item Let $\theta$ be any of the previous labellings.
\item Let $\spk_i$ be the i-vectors belonging to the speaker $i$.
\item Let $\Ymatd$ be the speaker identity variables of the
  development set. We will have as many identity variables as speakers.
\item Let $\Ymatt$ be the speaker identity variables of the test set.
\item Let $\Ymat$ be any of the previous speaker identity variables
  sets.
\item Let $\Xmatd$ be the channel variables of the
  development set. 
\item Let $\Xmatt$ be the channel variables of the test set.
\item Let $\Xmat$ be any of the previous channel variables
  sets.
\item Let
  $\Xmat_i=\left[\xvec_{i1},\xvec_{i2},\dots,\xvec_{iH_{i}}\right]$
  be the channel variables of speaker $i$.
\item Let $\model=\left(\muvec,\Vmat,\Umat,\Dmat\right)$ be the set of all
  the model parameters.
\end{itemize}

\section{Likelihood calculations}

\subsection{Definitions}

We define the sufficient statistics for speaker $i$. The zero-order
statistic is the number of observations of speaker $i$
$N_i$. The first-order and second-order statistics are
\begin{align}
  \Fvec_i=&\sumjHi\sumlLij \phivec_{ijl}\\
  \Smat_i=&\sumjHi\sumlLij \scatt{\phivec_{ijl}}
\end{align} 
We define the centered statistics as
\begin{align}
  \Fbar_i=&\Fvec_i-N_i \muvec\\
  \Sbarmat_i=&\sumjHi\sumlLij \scattp{\phivec_{ijl}-\muvec}
  =\Smat_i-\muvec\Fvec_i^T-\Fvec_i\muvec^T+N_i\scatt{\muvec}
\end{align}

We define the session statistics as
\begin{align}
  \Fvec_{ij}=&\sumlLij \phivec_{ijl} \\
  \Fbar_{ij}=&\Fvec_{ij}-L_{ij} \muvec
\end{align}
where $L_{ij}$ is the number of channels for the conversation $ij$.

We define the global statistics 
\begin{align}
  N&=\sumiM N_i \\
  \Fvec&=\sumiM \Fvec_i \\
  \Fbar&=\sumiM \Fbar_i \\
  \Smat&=\sumiM \Smat_i\\
  \Sbarmat&=\sumiM \Sbarmat_i
\end{align}

\subsection{Data conditional likelihood}

The likelihood of the data given the hidden variables for speaker $i$
is
\begin{align}
  \label{eq:plda_cond1}
  \lnProb{\spk_i|\yvec_i,\Xmat_i,\model}=&\sumjHi \sumlLij \ln
  \Gauss{\phivec_{ijl}}{\muvec+\Vmat\yvec_i+\Umat\xvec_{ij}}{\iWmat} \\
  \label{eq:plda_cond2}
  =&\frac{N_i}{2}\lndet{\frac{\Wmat}{2\pi}}
  -\med\sumjHi\sumlLij\mahP{\phivec_{ijl}}{\muvec-\Vmat\yvec_i
    -\Umat\xvec_{ij}}{\Wmat}\\
  \label{eq:plda_cond3}
  =&\frac{N_i}{2}\lndet{\frac{\Wmat}{2\pi}}-\med\trace\left(\Wmat\Sbarmat_i\right)
  +\yvec_i^{T}\Vmat^{T}\Wmat\Fbar_i-\frac{N_i}{2}\yvec_i^T\Vmat^T\Wmat\Vmat\yvec_i
  \nonumber \\
  &+\sumjHi
  \xvec_{ij}^{T}\Umat^{T}\Wmat\Fbar_{ij}
  -L_{ij}\yvec_i^{T}\Vmat^{T}\Wmat\Umat\xvec_{ij}
  -\med L_{ij}\xvec_{ij}^{T}\Umat^{T}\Wmat\Umat\xvec_{ij}
\end{align}

We can write this likelihood in other form if we define:
\begin{align}
  \ytildevec_{ij}&=
  \begin{bmatrix}
    \yvec_i\\
    \xvec_{ij} \\
    1
  \end{bmatrix}
  , & \quad \Vtildemat=
  \begin{bmatrix}
    \Vmat & \Umat & \muvec
  \end{bmatrix}
\end{align}
Then
\begin{align}
  \label{eq:plda_cond4}
  \lnProb{\spk_i|\yvec_i,\Xmat_i,\model}=&\sumjHi \sumlLij \ln
  \Gauss{\phivec_{ijl}}{\Vtildemat\ytildevec_{ij}}{\iWmat} \\
  \label{eq:plda_cond5}
  =&\frac{N_i}{2}\lndet{\frac{\Wmat}{2\pi}}
  -\med\sumjHi\sumlLij\mahP{\phivec_{ijl}}{\Vtildemat\ytildevec_{ij}}{\Wmat}
  \\
  \label{eq:plda_cond6}
  =&\frac{N_i}{2}\lndet{\frac{\Wmat}{2\pi}}-\med\trace\left(\Wmat\Smat_i\right)
  +\sumjHi \ytildevec_{ij}^{T}\Vtildemat^{T}\Wmat\Fvec_{ij}
  -\med L_{ij}\ytildevec_{ij}^T\Vtildemat^T\Wmat\Vtildemat\ytildevec_{ij}\\
  \label{eq:plda_cond7}
  =&\frac{N_i}{2}\lndet{\frac{\Wmat}{2\pi}}-\med\trace\left(\Wmat\left(\Smat_i+
      \sumjHi -2\Fvec_{ij}\ytildevec_{ij}^{T}\Vtildemat^{T}
      +L_{ij}\Vtildemat\ytildevec_{ij}\ytildevec_{ij}^T\Vtildemat^T\right)\right)
\end{align}

\subsection{Posterior of the hidden variables}

The posterior of the hidden variables can be decomposed into two
factors:
\begin{align}
  \label{eq:plda_xypost1}
  \Prob{\yvec_i,\Xmat_i|\Phimat_i,\model}=
  \Prob{\Xmat_i|\yvec_i,\Phimat_i,\model}\Prob{\yvec_i|\Phimat_i,\model}
\end{align}

\subsubsection{Conditional posterior of $\Xmat_i$}

The conditional posterior of $\Xmat_i$ is
\begin{align}
  \Prob{\Xmat_i|\yvec_i,\Phimat_i,\model}=
  \frac{\Prob{\Phimat_i|\yvec_i,\Xmat_i,\model}\Prob{\Xmat_i}}
  {\Prob{\Phimat_i|\yvec_i,\model}}
\end{align}

Using equations~\eqref{eq:plda_xprior} and~\eqref{eq:plda_cond3}
\begin{align}
  \label{eq:plda_xpost1}
  \lnProb{\Xmat_i|\yvec_i,\Phimat_i,\model}=&
  \lnProb{\Phimat_i|\yvec_i,\Xmat_i,\model}+\lnProb{\Xmat_i|\model}+\const\\
  \label{eq:plda_xpost2}
  =&\sumjHi
  \xvec_{ij}^{T}\Umat^{T}\Wmat\Fbar_{ij}
  -L_{ij}\yvec^{T}\Vmat^{T}\Wmat\Umat\xvec_{ij}
  -\med L_{ij} \xvec_{ij}^{T}\Umat^{T}\Wmat\Umat\xvec_{ij}
  -\med\xvec_{ij}^{T}\xvec_{ij}
  + \const\\
  \label{eq:plda_xpost2}
  =&\sumjHi
  \xvec_{ij}^{T}\Umat^{T}\Wmat\left(\Fbar_{ij}-L_{ij} \Vmat\yvec_i\right)
  -\med\xvec_{ij}^{T}\Lmatxij\xvec_{ij}+\const\\
  \label{eq:plda_xpost3}
  =&\sumjHi
  \xvec_{ij}^{T}\zetavec_{ij}
  -\med\xvec_{ij}^{T}\Lmatxij\xvec_{ij} + \const
\end{align}
where
\begin{align}
  \zetavec_{ij}=&\Umat^{T}\Wmat\left(\Fbar_{ij}
    -L_{ij}\Vmat\yvec_i\right)=\zetatildevec_{ij}-L_{ij}\Jmat\yvec_i\\
  \zetatildevec_{ij}=&\Umat^{T}\Wmat\Fbar_{ij}\\
  \Jmat=&\Umat^{T}\Wmat\Vmat\\
  \Lmatxij=&\Imat+L_{ij}\Umat^{T}\Wmat\Umat
\end{align}
Equation~\eqref{eq:plda_xpost3} has the form of a product of
Gaussian distributions. Therefore
\begin{align}
  \label{eq:plda_xpost4}
  \Prob{\Xmat_i|\yvec_i,\Phimat_i,\model}=
  \prod_{j=1}^{H_i}\Gauss{\xvec_{ij}}{\xbarvec_{ij}}{\iLmatxij}
\end{align}
where
\begin{align}
  \label{eq:plda_xpost5}
  \xbarvec_{ij}=\iLmatxij\zetavec_{ij}
\end{align}

\subsubsection{Posterior of $\yvec_i$}
The marginal posterior of $\yvec$ is
\begin{align}
  \label{eq:plda_ypost1}
  \Prob{\yvec|\spk_i,\model}=
  \frac{\Prob{\spk_i|\yvec_i,\model}\Prob{\yvec}}{\Prob{\spk_i|\model}}
\end{align}
We can use Bayes Theorem to write
\begin{align}
  \label{eq:plda_ypost2}
  \Prob{\spk_i,\Xmat_i|\yvec_i,\model}=
  \Prob{\spk_i|\yvec_i,\Xmat_i,\model}\Prob{\Xmat_i|\yvec_i,\model}
  =\Prob{\Xmat_i|\spk_i,\yvec_i,\model}\Prob{\spk_i|\yvec_i,\model}
\end{align}
Simplifying
\begin{align}
  \label{eq:plda_ypost3}
  \Prob{\spk_i|\yvec_i,\Xmat_i,\model}\Prob{\Xmat_i}=
  \Prob{\Xmat_i|\spk_i,\yvec_i,\model}\Prob{\spk_i|\yvec_i,\model}
\end{align}
Then
\begin{align}
  \label{eq:plda_ypost4}
  \Prob{\yvec|\spk_i,\model}=\left.\frac{\Prob{\spk_i|\yvec_i,\Xmat_i,\model}
      \Prob{\Xmat_i}\Prob{\yvec}}
    {\Prob{\Xmat_i|\spk_i,\yvec_i,\model}\Prob{\spk_i|\model}}\right|_{\Xmat_i=\zerovec}
\end{align}

Using equations~\eqref{eq:plda_yprior},~\eqref{eq:plda_cond3}
and~\eqref{eq:plda_xpost4}

\begin{align}
  \label{eq:plda_ypost5}
  \lnProb{\yvec_i|\spk_i,\model}=&\lnProb{\spk_i|\yvec_i,\Xmat_i,\model}+\lnProb{\yvec}
  -\lnProb{\Xmat_i|\spk_i,\yvec_i,\model}+\const\\
  \label{eq:plda_ypost6}
  =&\yvec^{T}\Vmat^{T}\Wmat\Fbar_i-
  \frac{N_i}{2}\yvec_i^T\Vmat^T\Wmat\Vmat\yvec_i-\med\yvec_i^T\yvec_i
  +\med\sumjHi\xAx{\xbarvec_{ij}}{\Lmatxij}+\const\\
  \label{eq:plda_ypost7}
  =&\yvec^{T}\Vmat^{T}\Wmat\Fbar_i
  -\med\yvec_i^T\left(\Imat+N_i\Vmat^T\Wmat\Vmat\right)\yvec_i\nonumber\\
  &+\med\sumjHi\xAxp{\Fbar_{ij}-L_{ij}\Vmat\yvec_i}{\Wmat\Umat\iLmatxij\Umat^{T}\Wmat}
  +\const\\
  \label{eq:plda_ypost8}
  =&\yvec^{T}\Vmat^{T}\Wmat\Fbar_i
  -\med\yvec_i^T\left(\Imat+N_i\Vmat^T\Wmat\Vmat\right)\yvec_i\nonumber\\
  &+\med\sumjHi\xAx{\Fbar_{ij}}{\Wmat\Umat\iLmatxij\Umat^{T}\Wmat} \nonumber\\
  &-2 L_{ij} \yvec_i^{T}\Vmat^{T}\Wmat\Umat\iLmatxij\Umat^{T}\Wmat\Fbar_{ij}
  \nonumber\\
  &+L_{ij}^2\xAx{\yvec_i}{\Vmat^{T}\Wmat\Umat\iLmatxij\Umat^{T}\Wmat\Vmat}
  +\const\\
  \label{eq:plda_ypost9}
  =&\yvec^{T}\Vmat^{T}\left(\Wmat\Fbar_i
    -\sumjHi L_{ij} \Wmat\Umat\iLmatxij\Umat^{T}\Wmat\Fbar_{ij}\right)\nonumber\\
  &-\med\yvec_i^T\left(\Imat+\Vmat^T
    \left(N_i\Wmat-\sumjHi
      L_{ij}^2\Wmat\Umat\iLmatxij\Umat^{T}\Wmat\right)
    \Vmat\right)\yvec_i
  +\const
\end{align}
Then
\begin{align}
  \label{eq:plda_ypost10}
  \Prob{\yvec_i|\spk_i,\model}=\Gauss{\yvec_i}{\ybarvec_i}{\iLmatyi}
\end{align}
where
\begin{align}
  \Lmatyi=&\Imat+\Vmat^T
  \left(N_i\Wmat-\sumjHi
    L_{ij}^2\Wmat\Umat\iLmatxij\Umat^{T}\Wmat\right)\Vmat\\
  =&\Imat+
  N_i\Vmat^T\Wmat\Vmat-\sumjHi L_{ij}^2\Jmat^{T}\iLmatxij\Jmat\\
  \gammvec_i=&\Vmat^{T}\left(\Wmat\Fbar_i
    -\sumjHi L_{ij} \Wmat\Umat\iLmatxij\Umat^{T}\Wmat\Fbar_{ij}\right)
  =\gammtildevec_i-\sumjHi L_{ij} \Jmat^{T}\iLmatxij\zetatildevec_{ij}\\
  \gammtildevec_i=&\Vmat^{T}\Wmat\Fbar_i\\
  \ybarvec_i=&\iLmatyi\gammvec_i
\end{align}

\subsection{Marginal likelihood of the data}

The marginal likelihood of the data is
\begin{align}
  \Prob{\spk_i|\model}=\left.\frac{\Prob{\spk_i|\yvec_i,\Xmat_i,\model}
      \Prob{\yvec_i}\Prob{\Xmat_i}}
    {\Prob{\Xmat_i|\yvec_i,\spk_i,\model}\Prob{\yvec_i|\spk_i,\model}}
  \right|_{\yvec_i=\zerovec,\Xmat_i=\zerovec}
\end{align}

Taking equations~\eqref{eq:plda_cond3}, \eqref{eq:plda_yprior},
\eqref{eq:plda_xprior} ~\eqref{eq:plda_xpost4}
and~\eqref{eq:plda_ypost10}
\begin{align}
  \label{eq:plda_pxi}
  \lnProb{\spk_i|\model}=\frac{N_i}{2}\lndet{\frac{\Wmat}{2\pi}}
  -\med\trace\left(\Wmat\Sbarmat_i\right)
  -\med\sumjHi\lndet{\Lmatxij}+\med\sumjHi\xAx{\zetatildevec_{ij}}{\iLmatxij}
  -\med\lndet{\Lmatyi}+\med\xAx{\gammvec_i}{\iLmatyi}
\end{align}

\section{EM algorithm}

\subsection{E-step}
In the E-step we calculate the posterior of $\yvec$ and $\Xmat$ with
equation~\eqref{eq:plda_xypost1} 

\subsection{M-step ML}

We maximize the EM auxiliary function $\Qcal(\model)$
\begin{align}
  \Qcal(\model)=&\sumiM
  \ExpcondYX{\ln\Prob{\spk_i,\yvec_i,\Xmat_i|\model}}\\
  =&\sumiM
  \ExpcondYX{\ln\Prob{\spk_i|\yvec_i,\Xmat_i,\model}}
  +\ExpcondYX{\lnProb{\yvec_i}}+\ExpcondYX{\lnProb{\Xmat_i}}
\end{align}
Taking equation~\eqref{eq:plda_cond7}
\begin{align}
  \Qcal(\model)=&
  \frac{N}{2}\lndet{\frac{\Wmat}{2\pi}}-\med\trace\left(\Wmat\left(\Smat+
      \sumiM \sumjHi -2\Fvec_{ij}\ExpcondYX{\ytildevec_{ij}}^{T}\Vtildemat^{T}
      +L_{ij} \Vtildemat\ExpcondYX{\ytildevec_{ij}\ytildevec_{ij}^T}
      \Vtildemat^T\right)\right)
\end{align}
we define
\begin{align}
  \Rmat_{\ytildevec}=&\sumiM \sumjHi L_{ij} \ExpcondY{\scatt{\ytildevec_{ij}}}\\
  \Cmat=&\sumiM \sumjHi \Fvec_{ij} \ExpcondY{\ytildevec_{ij}}^T
\end{align}
then
\begin{align}
  \Qcal(\model)=&\frac{N}{2}\lndet{\Wmat}-\med\trace\left(\Wmat\left(\Smat-2
      \Cmat\Vtildemat^{T}+
      \Vtildemat\Rmat_{\ytildevec}\Vtildemat^T\right)\right)+\const
\end{align}

\begin{align}
  \frac{\partial\Qcal(\model)}{\partial\Vtildemat}&=
  \Cmat-\Vtildemat\Rmat_{\ytildevec}=\zerovec \quad \implies\\
  \Vtildemat&=\Cmat\Rmat_{\ytildevec}^{-1}
\end{align}

\begin{align}
  \frac{\partial\Qcal(\model)}{\partial\Wmat}=\frac{N}{2}\left(2\Wmat^{-1}-\mathrm{diag}(\Wmat^{-1})\right)-\med\left(\Kmat+\Kmat^T-\mathrm{diag}(\Kmat)\right)=\zerovec
\end{align}
where $\Kmat=\Smat-2
\Cmat\Vtildemat^{T}+
\Vtildemat\Rmat_{\ytildevec}\Vtildemat^T$, so

\begin{align}
  \iWmat=&\frac{1}{N}\frac{\Kmat+\Kmat^T}{2}\\
  =&\frac{1}{N}\left(\Smat_{\phivec}-\Vtildemat\Cmat^T
    -\Cmat\Vtildemat^T+\Vtildemat\Rmat_{\ytildevec}\Vtildemat^T\right)\\
  =&\frac{1}{N}\left(\Smat-\Vtildemat\Cmat^T\right)
\end{align}

Finally, we need to evaluate the expectations $\ExpcondY{\ytildevec_{ij}}$
and $\ExpcondY{\scatt{\ytildevec_{ij}}}$ and compute
$\Rmat_{\ytildevec}$ and $\Cmat$.

\begin{align}
  \Cmat=\sumiM \sumjHi \Fvec_{ij} \ExpcondYX{\ytildevec_{ij}}^{T}
  =\sumiM \sumjHi \Fvec_{ij}
  \begin{bmatrix}
    \ExpcondY{\yvec_{i}}\\
    \ExpcondYX{\xvec_{ij}}\\
    1
  \end{bmatrix}^{T}=
  \begin{bmatrix}
    \Cmaty & \Cmatx & \Fvec
  \end{bmatrix}
\end{align}
Now

\begin{align}
  \ExpcondY{\yvec_i}=&\ybarvec_i\\
  \ExpcondYX{\xvec_{ij}}=&\ExpcondY{\xbarvec_{ij}}=
  \iLmatxij\left(\zetatildevec_{ij}-L_{ij}\Jmat\ybarvec_i\right)
\end{align}

\begin{align}
  \Cmaty=&\sumiM \sumjHi \Fvec_{ij} \ybarvec_i^{T}=\sumiM \Fvec_i
  \ybarvec_i^{T}\\
  \Cmatx=&\sumiM \sumjHi \Fvec_{ij}
  \left(\zetatildevec_{ij}-L_{ij}\Jmat\ybarvec_i\right)^{T} \iLmatxij
\end{align}

\begin{align}
  \Rmat_{\ytildevec}=&
  \begin{bmatrix}
    \Rmaty & \Rmatyx &
    \Rmatyo \\
    \Rmatxy & \Rmatx &
    \Rmatxo \\
    \Rmatyo^{T} &  \Rmatxo^{T} & N
  \end{bmatrix}
\end{align}

Now
\begin{align}
  \Rmatyo=&\sumiM N_{i} \ExpcondY{\yvec_i}=\sumiM N_i \ybarvec_i\\
  \Rmatxo=&\sumiM \sumjHi L_{ij} \ExpcondYX{\xvec_{ij}}=
  \sumiM \sumjHi L_{ij}\iLmatxij \left(
    \Umat^{T}\Wmat\Fbar_{ij}-L_{ij}\Jmat\ybarvec_i\right)\\
  \Rmaty=&\sumiM N_{i}\ExpcondY{\scatt{\yvec_{i}}}=\sumiM
  N_{i}\left(\iLmatyi+\scatt{\ybarvec_i}\right)\\
  \Rmatxy=&\sumiM\sumjHi L_{ij}\ExpcondYX{\xvec_{ij}\yvec_{i}^{T}}
  =\sumiM\sumjHi L_{ij}
  \ExpcondY{\iLmatxij\left(\zetatildevec_{ij}-L_{ij}\Jmat\yvec_i\right)\yvec_{i}^{T}}\\
  =&\sumiM \sumjHi L_{ij} \iLmatxij\left(\Umat^{T}\Wmat \Fbar_{ij} \ybarvec_i^{T}
    -L_{ij}\Jmat \ExpcondY{\scatt{\yvec_{i}}}\right)\\
  \Rmatx=&\sumiM \sumjHi L_{ij}\ExpcondYX{\scatt{\xvec_{ij}}}\\
  =&\sumiM \sumjHi L_{ij} \left(\iLmatxij+
    \iLmatxij
    \ExpcondY{\scattp{\Umat^{T}\Wmat\Fbar_{ij}-L_{ij}\Jmat\yvec_i}}
    \iLmatxij\right)\\
  =&\sumiM \sumjHi L_{ij} \left(\iLmatxij+
    \iLmatxij
    \left(\Umat^{T}\Wmat\Fbar_{ij}\Fbar_{ij}^T\Wmat\Umat \right.\right. \nonumber\\
  &\left.-L_{ij}\Umat^{T}\Wmat\Fbar_{ij}\ybarvec_i^{T}\Jmat^{T} 
    -L_{ij}\Jmat\ybarvec_i\Fbar_{ij}^T\Wmat\Umat \right. \nonumber\\
  &\left.\left.+L_{ij}^2\Jmat\ExpcondY{\scatt{\yvec_i}}\Jmat^{T}\right)\iLmatxij\right)
\end{align}

\subsection{M-step MD}

We assume a more general prior for the hidden variables:
\begin{align}
  \Prob{\yvec_i}=&\Gauss{\yvec_i}{\muvecy}{\iLambmaty}\\
  \Prob{\xvec_{ij}|\yvec_i}=&\Gauss{\xvec_{ij}}{\Hmat\yvec_i+\muvecx}{\iLambmatx}
\end{align}
To minimize the divergence we maximize
\begin{align}
  \Qcal(\muvecy,\Lambmaty,\Hmat,\muvecx,\Lambmatx)=&\sumiM \ExpcondY{\ln
    \Gauss{\yvec_i}{\muvecy}{\iLambmaty}}
  +\sumjHi\ExpcondYX{\ln\Gauss{\xvec_{ij}}{\Hmat\yvec_i+\muvecx}{\iLambmatx}}\\
  =&\frac{M}{2}\lndet{\Lambmaty}
  -\med\trace\left(\Lambmaty\sumiM \ExpcondY{\scattp{\yvec_i-\muvecy}}\right)\nonumber\\
  &+\frac{H}{2}\lndet{\Lambmatx}
  -\med\trace\left(\Lambmatx\sumiM\sumjHi
    \ExpcondYX{\scattp{\xvec_{ij}-\Hmat\yvec_i-\muvecx}}\right)\nonumber\\
  &+\const
\end{align}

\begin{align}
  \frac{\partial\Qcal(\muvecy,\Lambmaty,\Hmat,\muvecx,\Lambmatx)}
  {\partial\muvecy}&=\med\sumiM
  \Lambmaty\ExpcondY{\yvec_i-\muvecy}=\zerovec \quad \implies\\
  \muvecy&=\frac{1}{M}\sumiM\ExpcondY{\yvec_i}
\end{align}

\begin{align}
  \frac{\partial\Qcal(\muvecy,\Lambmaty,\Hmat,\muvecx,\Lambmatx)}
  {\partial\Lambmaty}&=
  \frac{M}{2}\left(2\Lambmaty^{-1}-\mathrm{diag}
    (\Lambmaty^{-1})\right)-\med\left(2\Smat-\mathrm{diag}(\Smat)\right)=\zerovec
\end{align}
where $\Smat=\sumiM\ExpcondY{\scattp{\yvec_i-\muvecy}}$, so

\begin{align}
  \Sigmaty=&\Lambmaty^{-1}\\
  =&\frac{1}{M}\sumiM\ExpcondY{\scattp{\yvec_i-\muvecy}}\\
  =&\frac{1}{M}\sumiM\ExpcondY{\scatt{\yvec_i}}
  -\muvecy\ExpcondY{\yvec_i}^{T}-\ExpcondY{\yvec_i}\muvecy^{T}
  +\scatt{\muvecy}\\
  =&\frac{1}{M}\sumiM\ExpcondY{\scatt{\yvec_i}}-\scatt{\muvecy}
\end{align}

\begin{align}
  \frac{\partial\Qcal(\muvecy,\Lambmaty,\Hmat,\muvecx,\Lambmatx)}{\partial\muvecx}=&
  \Lambmatx\sumiM\sumjHi
  \ExpcondYX{\xvec_{ij}-\Hmat\yvec_i-\muvecx}=\zerovec \quad \implies\\
  \muvecx=&\frac{1}{H}\left(\sumiM\sumjHi\ExpcondX{\xvec_{ij}}
    -\Hmat\sumiM H_i\ExpcondY{\yvec_i}\right)\\
  =&\frac{1}{H}\left(\Rhomatxo-\Hmat\Rhomatyo\right)
\end{align}
where
\begin{align}
  \Rhomatxo=&\sumiM\sumjHi\ExpcondX{\xvec_{ij}}\\
  \Rhomatyo=&\sumiM H_i\ExpcondY{\yvec_i}
\end{align}

\begin{align}
  \frac{\partial\Qcal(\muvecy,\Lambmaty,\Hmat,\muvecx,\Lambmatx)}{\partial\Hmat}=&
  \Lambmatx\sumiM\sumjHi
  \ExpcondYX{\left(\xvec_{ij}-\Hmat\yvec_i-\muvecx\right)\yvec_i^{T}}=\zerovec\\
  \implies&\Rhomatxy-\Hmat\Rhomaty-\muvecx\Rhomatoy=\zerovec\\
  \implies&\Rhomatxy-\Hmat\Rhomaty
  -\frac{1}{H}\left(\Rhomatxo-\Hmat\Rhomatyo\right)\Rhomat_{1y}=\zerovec\\
  \implies&\Rhomatxy-\frac{1}{H}\Rhomatxo\Rhomatoy
  -\Hmat\left(\Rhomaty-\frac{1}{H}\Rhomatyo\Rhomatoy\right)=\zerovec\quad\implies\\
  \Hmat=&\left(\Rhomatxy-\frac{1}{H}\Rhomatxo\Rhomatoy\right)
  \left(\Rhomaty-\frac{1}{H}\Rhomatyo\Rhomatoy\right)^{-1}
\end{align}
where
\begin{align}
  \Rhomatxy=&\sumiM\sumjHi\ExpcondYX{\xvec_{ij}\yvec_i^T}\\
  \Rhomaty=&\sumiM H_i \ExpcondY{\scatt{\yvec_i}} \\
  \Rhomatx=&\sumiM\sumjHi\ExpcondX{\scatt{\xvec_{ij}}}
\end{align}

\begin{align}
  \frac{\partial\Qcal(\muvecy,\Lambmaty,\Hmat,\muvecx,\Lambmatx)}
  {\partial\Lambmatx}&=
  \frac{H}{2}\left(2\Lambmatx^{-1}-\mathrm{diag}(\Lambmatx^{-1})\right)
  -\med\left(2\Smat-\mathrm{diag}(\Smat)\right)=\zerovec
\end{align}
where
$\Smat=\sumiM\sumjHi\ExpcondYX{\scattp{\xvec_{ij}-\Hmat\yvec_i-\muvecx}}$, so

\begin{align}
  \Sigmatx=&\Lambmatx^{-1}\\
  =&\frac{1}{H}\left(\Rhomatx
    -\Rhomatxy\Hmat^{T}-\Hmat\Rhomatxy^{T}
    -\Rhomatxo\muvecx^T-\muvecx\Rhomatxo^T
    +\Hmat\Rhomaty\Hmat^{T}\right.\nonumber\\
  &\left.+\Hmat\Rhomatyo\muvecx^T+\muvecx\Rhomatyo^T\Hmat^T
    +H\scatt{\muvecx}\right)\\
  =&\frac{1}{H}\left(\Rhomatx
    -\Rhomatxy\Hmat^{T}-\Hmat\Rhomatxy^{T}
    +\Hmat\Rhomaty\Hmat^{T}\right.\nonumber\\
  &\left.-\left(\Rhomatxo-\Hmat\Rhomatyo\right)\muvecx^T
    -\muvecx\left(\Rhomatxo-\Hmat\Rhomatyo\right)^T
    +H\scatt{\muvecx}\right)\\
  =&\frac{1}{H}\left(\Rhomatx
    -\Rhomatxy\Hmat^{T}-\Hmat\Rhomatxy^{T}
    +\Hmat\Rhomaty\Hmat^{T}
    -\left(\Rhomatxo-\Hmat\Rhomatyo\right)\muvecx^T\right)
\end{align}

The transform $(\yvec,\xvec)=\phi(\yvec^\prime,\xvec^\prime)$ such as
$\yvec^\prime$ and $\xvec^\prime$ has a standard prior is
\begin{align}
  \yvec=&\muvecy+(\Sigmaty^{1/2})^{T}\yvec^\prime\\
  \xvec=&\muvecx+\Hmat\yvec+(\Sigmatx^{1/2})^{T}\xvec^\prime\\
  =&\muvecx+\Hmat\muvecy+\Hmat(\Sigmaty^{1/2})^{T}\yvec^\prime
  +(\Sigmatx^{1/2})^{T}\xvec^\prime
\end{align}

We can transform $\muvec$, $\Vmat$ and $\Umat$ using that transform
\begin{align}
  \Umat^\prime=&\Umat(\Sigmatx^{-1/2})^T\\
  \Vmat^\prime=&\left(\Vmat+\Umat\Hmat\right)(\Sigmaty^{-1/2})^T\\
  \muvec^\prime=&\muvec+\left(\Vmat+\Umat\Hmat\right)\muvecy+\Umat\muvecx
\end{align}

\subsection{Objective function}

The EM objective function is equation~\eqref{eq:plda_pxi} summed for
all speakers
\begin{align}
  \label{eq:plda_obj}
  \lnProb{\Phimat|\model}=&\frac{N}{2}\lndet{\frac{\Wmat}{2\pi}}
  -\med\trace\left(\Wmat\Sbarmat\right)
  -\med\sumiM\sumjHi\lndet{\Lmatxij}
  +\med\sumiM\sumjHi\xAx{\zetatildevec_{ij}}{\iLmatxij} \nonumber \\
  &-\med\sumiM\lndet{\Lmatyi}+\med\sumiM\xAx{\gammvec_i}{\iLmatyi}
\end{align}

\section{Likelihood ratio}

Given a model $\model$ we can calculate the ratio of the posterior
probabilities of target and non target as shown in \cite{Brummer2010}:
\begin{align}
  \label{eq:llk_plugin}
  \frac{\Prob{\tar|\Phimatt,\model,\pi}}{\Prob{\nontar|\Phimatt,\model,\pi}}=\frac{\Ptar}{\Pnon}\frac{\Prob{\Phimatt|\tar,\model}}{\Prob{\Phimatt|\nontar,\model}}=\frac{\Ptar}{\Pnon}\Rat{\Phimatt,\model}
\end{align}
where we have defined the plug-in likelihood ratio
$\Rat{\Phimatt,\model}$. To get this ratio we need to
calculate $\Prob{\Phimat|\theta,\model}$. Given a model $\model$, the
$\yvec_1,\yvec_2,\dots,\yvec_M \in \Ymat$ are sampled independently
from $\Prob{\yvec|\model}$. Besides, given the $\model$ and a speaker
$i$ the set $\spk_i$ of i-vectors produced by that speaker are drown
independently from $\Prob{\Phimat|\yvec_i,\model}$. Using these
independence assumptions we can write:
\begin{align}
  \label{eq:PPhi_thetaM}
  \Prob{\Phimat|\theta,\model}&=\prod_{i=1}^{M} \Prob{\spk_i|\model} \\
  \label{eq:Pspki_yM}
  \Prob{\spk_i|\yvec,\model}&=\prod_{\phivec \in \spk_i} \Prob{\phivec|\yvec,\model}
\end{align}
Then, the likelihood of $\Phimat$ is
\begin{align}
  \label{eq:PPhi_thetaM_2}
  \Prob{\Phimat|\theta,\model}=&\prod_{i=1}^{M}\frac{\Prob{\spk_i|\yvec_0,\model}\Prob{\yvec_0|\model}}{\Prob{\yvec_0|\spk_i,\model}}=K(\Phimat)L(\theta|\Phimat)
\end{align}
where $K(\Phimat)=\prod_{i=1}^{N} \Prob{\phivec_j|\yvec_0,\model}$ is
a term that only dependent on the dataset, not $\theta$, so it
vanishes when doing the ratio and we do not need to calculate it. What
we need to calculate is:
\begin{align}
  \label{eq:L_plugin}
  L(\theta|\Phimat)&=\prod_{i=1}^{M} \Q{\spk_i} \\
  \label{eq:Q_plugin}
  \Q{\spk_i}&=\frac{\Prob{\yvec_0|\model}}{\Prob{\yvec_0|\spk_i,\model}} 
\end{align}
and the likelihood ratio is:

\begin{equation}
  \Rat{\Phimatt,\model}=\frac{\Q{\left\{l,r\right\}}}{\Q{\left\{l\right\}}\Q{\left\{r\right\}}}
\end{equation}

Making $\yvec_0=0$ we can get use~\eqref{eq:plda_ypost4}, \eqref{eq:plda_yprior} to calculate
$\Q{\spk}$

\begin{equation}
  \lnQ{\spk_i}=\med\left(-\lndet{\Lmatyi}+\xAx{\gammvec_i}{\iLmatyi}\right)
\end{equation}

Given a set of training observations $\spk_1$ of a speaker 1 with
statistics $N_1$ and $\Fbar_1$; and
a set of test observations $\spk_2$ of a speaker 2 with statistics
$N_2$ and $\Fbar_2$. To test if the speakers
1 and 2 are the same speaker the log-likelihood ratio is
\begin{align}
  \ln \Rat{\Phimatt,\model}=\med\left(-\lndet{\Lmat_3}+\xAx{\gammvec_3}{\iLmat_3}+\lndet{\Lmat_1}-\xAx{\gammvec_1}{\iLmat_1}+\lndet{\Lmat_2}-\xAx{\gammvec_2}{\iLmat_2}\right)
\end{align}
where
\begin{align}
  \Prob{\yvec|\spk_1,\model}=&\Gauss{\yvec}{\gammvec_1\iLmat_1}{\iLmat_1}\\
  \Prob{\yvec|\spk_2,\model}=&\Gauss{\yvec}{\gammvec_2\iLmat_2}{\iLmat_2}\\
  \Prob{\yvec|\spk_1,\spk_2,\model}=&\Gauss{\yvec}{\gammvec_3\iLmat_3}{\iLmat_3}\\
\end{align}

Using that $\gammvec_3=\gammvec_1+\gammvec_2$:
\begin{align}
  \ln \Rat{\Phimatt,\model}=\med\left(\lndet{\Lmat_1}+\lndet{\Lmat_2}-\lndet{\Lmat_3}+2\gammvec_1^T\iLmat_3\gammvec_2+\xAx{\gammvec_1}{(\iLmat_3-\iLmat_1)}+\xAx{\gammvec_2}{(\iLmat_3-\iLmat_2)}\right)
\end{align}

\bibliographystyle{IEEEbib}
\bibliography{villalba}

\end{document}